\documentclass[twocolumn,longbibliography, superscriptaddress, english, prb]{revtex4-1}
\usepackage[colorlinks=true,urlcolor=blue,citecolor=blue,linkcolor=blue]{hyperref}
\usepackage[T1]{fontenc}
\usepackage[latin9]{inputenc}
\usepackage{amssymb}
\usepackage{graphicx}
\usepackage{amsmath,color}
\usepackage{mathrsfs}
\usepackage{float}
\usepackage{indentfirst}
\usepackage{braket}
\usepackage{comment}
\usepackage{txfonts}
\usepackage{booktabs}

\usepackage{algpseudocode} 
\usepackage{algorithm}

\def\journal #1, #2, #3, 1#4#5#6{{\sl #1~}{\bf #2}, #3 (1#4#5#6) }

\def\eqa{\begin{eqnarray}}
\def\eea{\end{eqnarray}}
\newcommand{\eq}{\begin{equation}}
\newcommand{\ee}{\end{equation}}

\renewcommand{\mathbf}[1]{\ensuremath{\boldsymbol{ #1}} }

\usepackage{babel}

\begin{document}
\title{Supervised Learning with Projected Entangled Pair States}
\author{Song Cheng}
\affiliation{Beijing Institute of Mathematical Sciences and Applications, Beijing 101407, China}
\affiliation{Institute of Physics, Chinese Academy of Sciences, Beijing 100190, China}
\author{Lei Wang}
\email{wanglei@iphy.ac.cn}
\affiliation{Institute of Physics, Chinese Academy of Sciences, Beijing 100190, China}
\affiliation{Songshan Lake Materials Laboratory, Dongguan, Guangdong 523808, China}

\author{Pan Zhang}
\email{panzhang@itp.ac.cn}
\affiliation{CAS key laboratory of theoretical physics, Institute of Theoretical Physics, Chinese Academy of Sciences, Beijing 100190, China}

\begin{abstract}
Tensor networks, a model that originated from quantum physics, has been gradually generalized as efficient models in machine learning in recent years.
However, in order to achieve exact contraction, only tree-like tensor networks such as the matrix product states and tree tensor networks have been considered, even for modeling two-dimensional data such as images.
In this work, we construct supervised learning models for images using the \textit{projected entangled pair states} (PEPS), a two-dimensional tensor network having a similar structure prior to natural images.
Our approach first performs a feature map, which transforms the image data to a product state on a grid, then contracts the product state to a PEPS with trainable parameters to predict image labels. The tensor elements of PEPS are trained by minimizing differences between training labels and predicted labels.
The proposed model is evaluated on image classifications using the MNIST and the Fashion-MNIST datasets. We show that our model is significantly superior to existing models using tree-like tensor networks. Moreover, using the same input features, our method performs as well as the multilayer perceptron classifier, but with much fewer parameters and is more stable. Our results shed light on potential applications of two-dimensional tensor network models in machine learning. 
\end{abstract}

\maketitle

\section{Introduction}

Tensor networks is a framework that approximates a high-order tensor using contractions of low-order tensors. It has been widely used in quantum many-body physics~\cite{Orus2014, orus_advances_2014,Verstraete:2008ko} whose correlation or entanglement entropy satisfies the area law~\cite{PhysRevLett.100.070502,ToricPeps,Eisert:2010hq}, which is traditionally related to the partition function of the local classical Hamiltonians and the gapped ground state of the local quantum Hamiltonians. Recent works have uncovered that the information entropy of image datasets in machine learning also approximately satisfies the area law~\cite{deepandcheaplearn,MI_localRBM}, which inspired several recent works introducing tensor networks to various machine learning tasks~\cite{2017arXiv170104844D,2017arXiv170105039G,PhysRevB.97.085104,MPSSL,stoudenmire2018learning,han2018unsupervised,Liu2017,glasser2018supervised}.

On the theoretical side, introducing tensor networks which is purely linear can bring more interpretability to the machine learning framework: the entanglement structure can be naturally introduced to characterize the expressive power of the learning model, and some neural network models can be mapped to a tensor network for theoretical investigations~\cite{PhysRevB.97.085104,levine2019quantum}. On the practical side, numerical techniques of the tensor networks are also useful and inspiring for optimizing and training methods of the machine learning models, these include the normalized (or canonical) form, the adaptive learning, etc.~\cite{MPSSL,han2018unsupervised,stoudenmire2018learning}. These benefits have motivated Google X of Alphabet to release a \textit{TensorNetwork} library~\cite{roberts2019tensornetwork} built upon Google's famous machine learning framework~\textit{tensorflow}~\cite{tensorflow2015}.

It has been shown, tensor networks has a deep connection with the probabilistic graph models in statistical learning, including the hidden Markov chain~\cite{MPSSL}, the restricted Boltzmann machine~\cite{PhysRevB.97.085104}, the Markov random fields~\cite{glasser2018supervised}, etc. The study of this connection has given rise to the so call "Born Machine"~\cite{MI_localRBM} as a typical quantum machine learning prototype. In this sense, the tensor networks based machine learning is closely related to the quantum machine learning in terms of their model structure. A special part of tensor networks based machine learning algorithm can be directly regarded as the classical simulation of the corresponded quantum machine learning algorithm~\cite{2018arXiv180404168L,gao_efficient_2017}. 

\begin{figure}
\begin{center}
\includegraphics[width=\columnwidth]{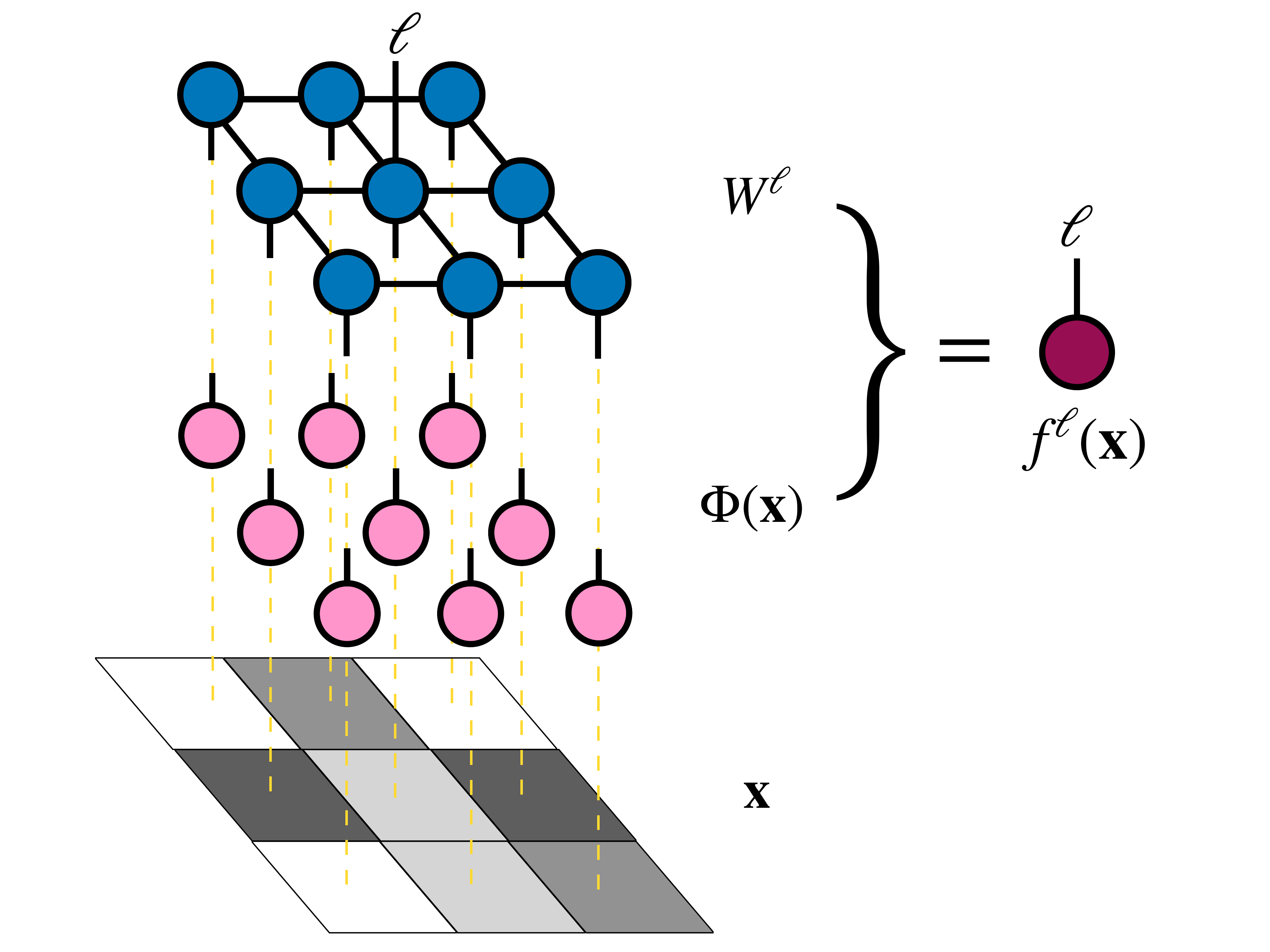}
\caption{Supervised learning model of PEPS structure. The input image $\mathbf{x}$ would be mapped to a high-dimensional vector $\Phi(\mathbf{x})$ consisting of local feature maps $\phi_{s_i}(\mathbf{x_i})$. The label vetor $f^\ell(\mathbf{x})$ come from the contraction of $\Phi(\mathbf{x})$ and a PEPS strucure tensor network $\mathbf W$. } 
\label{fig:weight}
\end{center}
\end{figure}

In recent years, various of tensor networks models, such as the \textit{Matrix Product States}(MPS)~\cite{MPSSL,han2018unsupervised}, the \textit{Tree Tensor Network}(TTN)~\cite{stoudenmire2018learning,Liu2017,Cheng2019,stoudenmire2018learning}, the \textit{String Bond States}(SBS)~\cite{glasser2018supervised}, have been introduced to machine learning, and found successful applications in image classifications, image density estimation, generative modeling of natural languages~\cite{guo2018matrix}, and neural network compressions~\cite{novikov2015tensorizing}. These models are all based on the quasi-one-dimensional tree-like tensor networks, which are the best-understood types of tensor networks, and can be efficiently contracted due to the existence of the canonical form. 
However, when dealing with data such as natural images, the spatial correlations between nearby pixels  as well as the structural prior are completely ignored in the tree-like tensor networks, forcing some short-range correlations to be artificially made long-range, leading to unnecessary computational overhead as well as statistical bias. 
Notice that in physics there is one kind of tensor network with exactly the same geometric structure as the natural images, known as \textit{Projected Entangled Pair States} (PEPS)~\cite{Orus2014,Verstraete:2008ko}. which is composed of tensors located on a two-dimensional lattice, as shown in Fig.~\ref{fig:weight}.

In this work, we introduce a supervised learning model based on the PEPS representations on a $L\times L$ grid with local physical dimension $d$ and local bond dimension $D$ for each tensor. The size of PEPS $L$ and value of local physical dimension $d$ depend on the feature map we employ, which transforms the set of $L_0\times L_0$ pixels of an input image to a product state on a $L\times L$ grid. Given the features, a PEPS with trainable parameters contracts with input feature to obtain a probability distribution over labels, then predict the label of the input image. When the grid is small, exact contractions of PEPS can be performed with both space and time complexity proportional to the $D^L$. When $L$ or $D$ is large, the PEPS can not be contracted exactly, then we apply an approximate contraction method based on the boundary MPS method. We evaluate the PEPS model on the standard MNIST and the Fashion-MNIST datasets. We show that our model significantly outperforms existing tensor-network models using MPS and tree tensor networks. When compared with the standard classifier, the Multilayer Perceptron (MLP), we find that they perform similarly when the same input features are used, but our PEPS based method requires much fewer parameters and is more stable.

The rest of this paper is organized as follows: In Sec.~\ref{sec:model} we give a detailed description of the PEPS model and the corresponding training algorithm. In Sec.~\ref{sec:experiment} we evaluate our model on the MNIST~\cite{MNIST} and the Fashion-MNIST~\cite{xiao2017fashion} datasets and compared the results with other tensor network models as well as classic machine learning models. We conclude in Sec.~\ref{sec:discussion} and discuss possible future developments along the direction of applying tensor networks to machine learning.

\section{Image classification with PEPS\label{sec:model}}

\subsection{Feature map of input data}

The goal of supervised learning is to learn a complex function $f(\mathbf{x})$ which maps an input training (grayscale) image $\mathbf{x}\in\mathbb R^{L_0\times L_0}$ with pixels defined on a $L_0\times L_0$ grid, to a given label $y\in \{1,2,...,T\}$, where $T$ denotes the number of possible labels. Usually, such mapping is highly nonlinear in the original space of input data $\mathbf{x}$, because nonlinearity effectively increases the dimension of the input space where features of data are easier to capture.
In this work, we consider the classifier with tensor networks, which is a linear model usually acting at a space with a very large dimension. The motivation of working with a very large dimension is that there is not necessary to consider nonlinearity because all features would become linear separable as stated in the representer theorem~\cite{scholkopf2001generalized}. So first one needs to transform the input data $\mathbf{x}$ to a \textit{feature tensor} $\Phi(\mathbf{x})$ in a space of large dimension using a feature map. We consider two distinct kinds of feature maps in this work.

\subsubsection{Product state feature map}
A simple way to increase the dimension of input space is creating an Hilbert space for pixels. This is to levarage the black pixel with $x_i=0$ and white pixel with $x_i=1$ as a black state $|0\rangle = \left ( \begin{matrix}1 \\ 0 \end{matrix}\right )$ and a white state $|1\rangle = \left ( \begin{matrix}0 \\ 1 \end{matrix}\right )$ respectively, then convert each gray scale pixel $x_i$ in the image $\mathbf x$ as a super position of $|0\rangle$ and $|1\rangle$
    \begin{equation}
\phi(x_i) = a\left ( \begin{matrix}1 \\ 0 \end{matrix}\right ) + b\left ( \begin{matrix}0 \\ 1 \end{matrix}\right ),
    \end{equation}
    where $a$ and $b$ are functions of $x_i$, which for example can be chosen as

    \begin{equation}
        a=\cos(\pi x_i/2),\,\,\,\,\,\,\,\,\,\,\,\,b=\sin(\pi x_i/2).
\label{eq:feature_map3}
    \end{equation}
    For image with $N=L_0\times L_0$ pixels, the feature tensor $\Phi(\mathbf{x})$ is then defined by the tensor product of $\phi(x_i)$
\begin{equation}
\Phi(\mathbf{x}) =  \phi(x_1)\otimes\phi(x_2)\otimes \cdots \otimes \phi(x_N)
\label{eq:feature_map2}
\end{equation}
This is probably the most straightforward feature map that transforms every pixel in the original space $\mathbb R^{N}$ to a product state in the Hilbert space of dimension $2^{L_0\times L_0}$, and has been widely used in the literatures~\cite{MPSSL}. We term it as the \textit{product state feature map}.

\subsubsection{Convolution feature map}
\label{sec:cnn}
The simple product state feature maps introduced in the previous section are pre-determined before the classifier is applied, thus is apparently not optimal. Another option is using an adaptive feature map with parameters learned together with the classifier. The most famous adaptive feature map is the convolution layers, which perform non-linear transformations to transform input images to a feature tensor with multiple channels ~\cite{Goodfellow-et-al-2016-Book} using two-dimensional convolutions.

The input of the convolution layer is a raw image $\mathbf x\in\mathbb R^{L_0\times L_0}$. After the transformation, the convolution layer outputs
a three-order feature tensor with dimension $L\times L\times d$,  where the $L\times L$ refers to the output size of features with $L\leq L_0$ depending on size kernels and paddings, and $d$ denotes the number of channels. This is to say that the output of the CNN feature map is also a product state with components located at a grid of size $L\times L$, and each component is of local physical dimension $d$. Thus the total space size of the feature tensor is $d^{L\times L}$.

In the standard convolution neural networks (CNN), the function of convolution layers (plus pooling layers) is extracting relevant features from input data. Following the convolution layers, a classifier, usually a multi-layer perceptron (MLP), or simply a linear classifier such as the Logistic regression, is used for predicting a label from the extracted features. Notice that the MLP can not accept a feature tensor as input. Instead, in MLP the feature tensor $\Phi(\mathbf x)$ is flattened to a vector in the space $\mathbb R^{d\times L\times L}$, that is, completely ignored the spatial structure of the feature tensor.
In this work, we consider a linear classifier using two-dimensional tensor networks, which can fully take the raw feature tensor as an input, as tensor networks was born to compress a large Hilbert space.

\subsection{PEPS Classifier}
Equipped with the feature map, and the extracted feature tensor $\Phi(\mathbf x)\in \mathbb R^{d^{L\times L}}$, we then consider a linear mapping $\mathbf W$, which results to a vector representing probablity of being one of $T$ labels given the input image.  To be more specific, consider
\begin{equation}
\mathbf{f}(\mathbf{x}) = \mathbf W \cdot \Phi(\mathbf{x})
\label{eq:kernel}
\end{equation}
Where $\mathbf W\in \mathbb R^{d^{L\times L}\times T}$, and $\cdot$ denotes tensor contraction of the operator $\mathbf W$ and feature tensor $\Phi(\mathbf x)$.
In general $\mathbf W$ is a tensor of order $L\times L+1$, where $L\times L$ is the order of input, each of which has dimension $d$, and the output has dimension $T$. We can see that the total number of parameters, $d^{L\times L}\times T$, is too large to use in practice. So one must make a reasonable approximation of $\mathbf W$ to archive a practical and efficient algorithm, or in other words, to decompose $\mathbf W$ into contraction of many adjacent tensors using the tensor network representation. 

For physics systems, such as the gapped ground states of the local quantum hamiltonian, most long-range correlations of the model are irrelevant, the tensor networks approach has been proved to be a successful ansatz.
For systems like the natural image datasets, its locality has been discussed in several works recently~\cite{MI_localRBM,2017arXiv171005520Z,deepandcheaplearn}. This suggests that the correlation of the natural images handled by the machine learning field may be dominated by local correlations, with rare long-range correlations. This implication has also been numerically verified by some successful tensor network machine learning models based on the MPS~\cite{MPSSL,han2018unsupervised}, TTN~\cite{Liu2017,stoudenmire2018learning}, etc. 
However, we noticed that previous works on compressing $\mathbf W$ are all based on tree-like tensor networks, which completely ignored the two-dimensional nature of feature tensor $\Phi(\mathbf x)$, treating it as a quasi-one-dimensional tensor. In this work, we propose to use the PEPS which is a two-dimensional tensor networks with a similar structural prior to images.

As shown in Fig.~\ref{fig:weight}, the PEPS represents the $\mathbf W$ using a composition of tensors $T^{[i]}$, 
\begin{equation}
W^{\ell,s_1s_2\cdots s_N} = \sum_{\sigma_1\sigma_2 \cdots \sigma_K} T^{s_1}_{\sigma_{1},\sigma_{2}}T^{s_2}_{\sigma_{3},\sigma_{4},\sigma_{5}} \cdots T^{s_i,\ell}_{\sigma_{k},\sigma_{k+1},\sigma_{k+2},\sigma_{k+3}} \cdots T^{s_N}_{\sigma_{K-1},\sigma_{K}}
\label{eq:peps}
\end{equation}
where $K$ is the number of bonds in the square lattice. Each tensor has a "physical" index $s_i\in\{1,2,...,d\}$ connected to the input vector $\Phi(x_{i})$. Their "virtual" indices $\sigma_k\in\{1,2,...,D\}$ will be contracted with their adjacent tensors of a square lattice. Besides, a special tensor in the center of the lattice has an additional "label" index $\ell\in \{1,2,...,T\}$ to generate the output vector of the model. The values of these tensors are randomly initialized by real numbers ranging from 0 to 0.01, which constitute the trainable parameters of the model, denoted as $\theta$ .

\subsection{Training algorithm}
The purpose of training is to minimize the difference between the predicted labels and the training labels by tuning the training parameters $\theta$. This is usually achived by minimizing a loss function $\mathcal L$ representing distance between distribution of predicted labels and the one-hot vector correponding to the distribution of training labels. A common choice of the loss function is defined as
\begin{equation}
    \mathcal{L} = -\sum_{\mathbf{x}_i, y_i \in \mathcal{T}}\log \left[ \mathrm{softmax} \left(f^{[y_i]}(\mathbf{x}_i)\right) \right ],
\label{eq:loss}
\end{equation}
where
\begin{equation}
\mathrm{softmax} [f^{[y_i]}(\mathbf{x})]  \equiv \frac{\exp{f^{[y_i]}(\mathbf{x}_i)}}{\sum_{\ell=1}^{T}\exp{f^{[\ell]}(\mathbf{x}_i)}}.
\end{equation}
Here $\mathbf{x}_i, y_i$ denotes the $i$ th image and the corresponding label in the dataset $\mathcal{T}$. The output of softmax function can be interpreted as the probability that the model predicts that $\mathbf{x}_i$ belongs to class $y_i$. $\mathcal{L}$ is the \textit{cross-entropy} of model's probability and image's labels, which is known to be well suited for most of supervised learning models. $f^\ell(\mathbf{x})$ comes from the contraction of the physical indices of the PEPS model and the feature map vectors.

\begin{equation}
\begin{split}
    f^{[\ell]}(\mathbf{x}) &= W^{\ell,s_1s_2\cdots s_N} \cdot \phi_{s_1}(x_1)\otimes\phi_{s_2}(x_2)\otimes \cdots \otimes \phi_{s_N}(x_N) \\
&= \sum_{\sigma_1\sigma_2 \cdots \sigma_K} M_{\sigma_{1},\sigma_{2}}M_{\sigma_{3},\sigma_{4},\sigma_{5}} \cdots M^{\ell}_{\sigma_{k},\sigma_{k+1},\sigma_{k+2},\sigma_{k+3}} \cdots M_{\sigma_{K-1},\sigma_{K}}
\end{split}
\label{eq:contraction}
\end{equation}
where each $M = \sum_{s_i}T^{s_i}\phi_{s_i}(x_i)$.

\begin{figure}
\begin{center}
\includegraphics[width=\columnwidth]{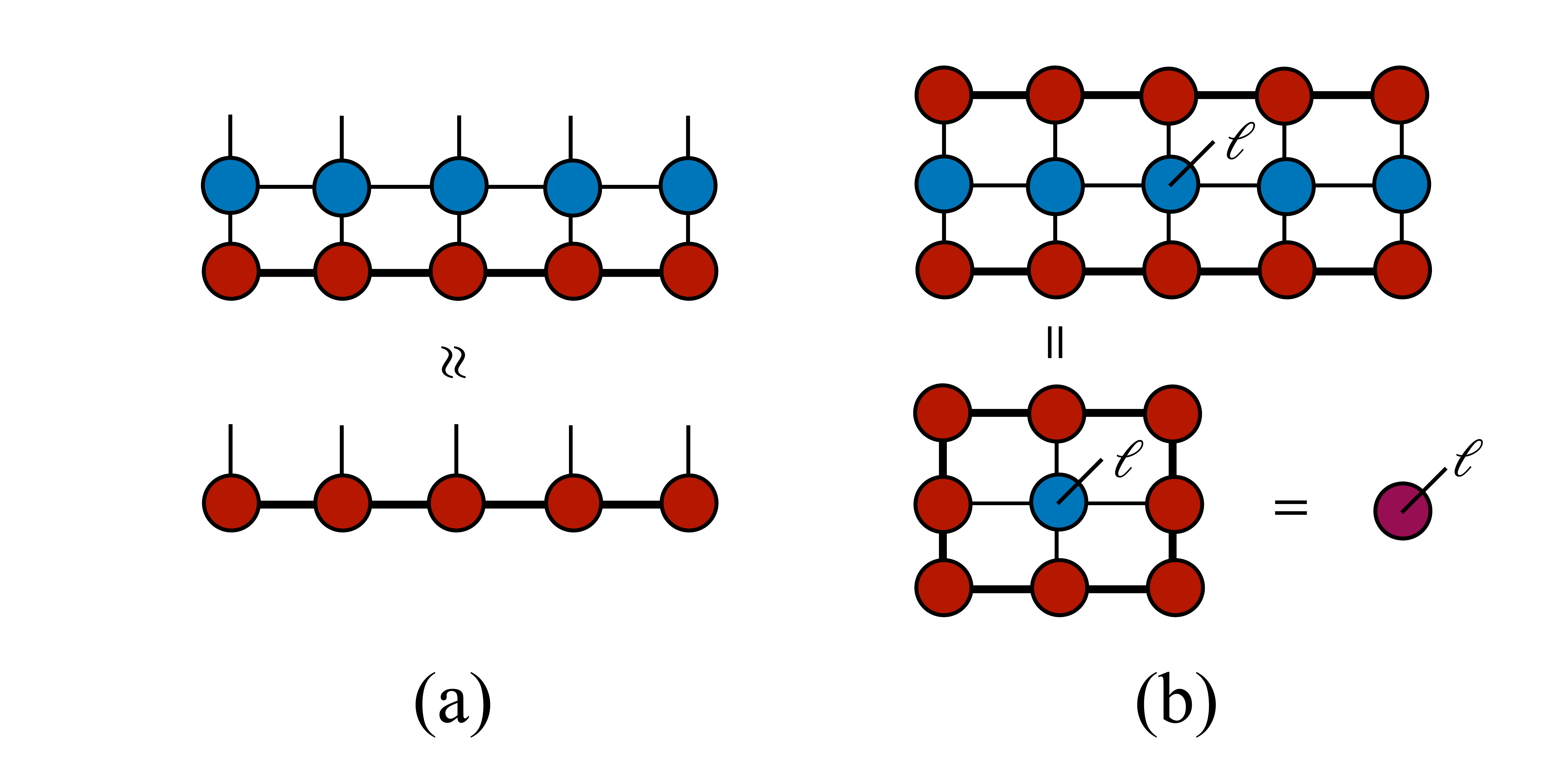}
\caption{\textbf{(a)} Contracting tensors of one row to another causes the bond dimension to grow exponentially. To keep $D$ under a finite number $\chi$, an approximate algorithm likes the Boundary MPS is inevitable. \textbf{(b)}. After several approximate contraction, the contractions of the last lines can be exactly performed under the complexity of $~O(T\chi^3D^2)$. \label{fig:contraction}}
\end{center}
\end{figure}

However, calculating $f^{\ell}(\mathbf{x}_i)$ from contracting these M tensors is not trivial.
Consider contracting the tensors of the bottom row with the tensors of the nearest row, as shown in Fig.~\ref{fig:contraction}~(a), if the initial dimension of the virtual bond of tensors is $D$, the bond dimension of the resulting tensors would increase from $D$ to $D^2$. As this process repeats, the computational cost of contraction would grow exponentially. More rigorous proof shows that the exact contraction of the PEPS structure is \#P hard~\cite{PhysRevLett.98.140506}, there is no polynomial algorithm exists in general.

If the PEPS is small, one can do the contraction with computational complexity proportional to $D^{L}$. If $D$ and lattice length $L$ are large, we have to use approximate methods. one of them is the Boundary MPS method~\cite{jordan2008classical}, which treats the bottom row tensors as an MPS and the rest of row tensors as the operators applied on the MPS. Each time the neighbor row tensors are applied, a DMRG-like method is used to truncate the bond dimension of the MPS to a maximum value $\chi$. In order to achieve a smaller truncation error, one first applies the QR decomposition to the MPS to ensure that the MPS is in the correct canonical form. Then apply the SVD decomposition to the central tensor of the canonicalized MPS, which ensures the truncation is optimal for the entire row. 
There are many choices for the approximate contraction methods, such as the coarse-grained methods~\cite{TRGLevin}, closely related to the renormalization group theory, which may provide a natural way for introducing the renormalization group into machine learning. However, those methods usually have a much larger computational complexity than the boundary MPS method, and are more suitable for infinite systems.

The total computational cost of the approximate contraction by the Boundary MPS is $\sim O(N\chi^3D^6)$. 
More efficiently, We can approximately contract from the top and the bottom in parallel. The resulting tensor network is shown in Fig.~\ref{fig:contraction}~(b), which can be contracted exactly with calculation complexity~$\sim O(T\chi^3D^2)$.
The computational cost of this training algorithm for a complete forward process scales as $\sim O(|\mathcal{T}|N\chi^3D^6)$, where $|\mathcal{T}|$ denotes the number of input images, usually $|\mathcal{T}| \sim 50000$, the cost of this algorithm will therefore be significantly larger than the corresponding algorithm in the many-body physics. 

In addition to the forward process for predicting the labels and evaluate the loss function, we also need the backward process to compute the gradients of the loss function with respect to the training parameters.
In this work, we adopt the automatic differentiation technique of tensor networks~\cite{PhysRevX.9.031041}. The key to this technique is to treat the tensor network algorithm, such as the Boundary MPS, as a traceable computation graph about tensors and algebraic operations. Then through simple chain rule, one could implement a backward-propagation process along this computation graph to get the gradient value of the loss function $\mathcal{L}$ with respect to each parameter. Automatic differentiation is one of the core techniques of modern machine learning applications. It is proven that the computational complexity would not exceed the original feed-forward algorithm.

There are two obstacles to applying automatic differentiation to tensor networks. The first is that in the backward process of singular value decompositions (SVD), in case of existing singular value degeneracies $\lambda_i = \lambda_j$, a factor appearing in the perturbation analysis $F_{ij} \equiv \frac{1}{\lambda_i - \lambda_j}$, would encounter numerical instability. 
The solution is to replace $F_{ij}$ with $\frac{\lambda_i - \lambda_j}{(\lambda_i - \lambda_j)^2+\epsilon}$, where $\epsilon$ is a small factor that does not significantly change the gradient~\cite{PhysRevX.9.031041}. 
The second issue is that automatic differentiation tensor networks may cause huge memory consumption. We employ two techniques to settle it. One is called \textit{checkpointing}~\cite{2016arXiv160406174C} technique, which stores less intermediate variables by recalculating them during the backward process. The other is \textit{blocking}, which uses one tensor whose physical index dimension is $d^n$ to parameterize $n$ neighboring pixels to reduce the scale of the PEPS, which is equivalent to approximating an intermediate tensor of bond dimension $D^{\frac{n}{2}}$ with a tensor of bond dimension $D$.

After obtaining the gradients, we directly apply the Stochastic gradient descent (SGD) and Adam optimizer~\cite{2014arXiv1412.6980K} to update the trainable variables $\theta$. However, we found that in practical situations, when the feature map is a simple function similar to Eq.~\ref{eq:feature_map2} other than CNN, if we keep the parameters non-negative, such as assigning $|\theta|$ to $\theta$, the stability of optimization can be greatly improved. Under this constraint, the PEPS model can also be regarded as a kind of Markov Random Field(MRF) of probabilistic graph models.
The reason for these phenomena may be attributed to the fact that the SVD truncation error of the positive matrices is usually smaller, so the gradient value could be transmitted more accurately. These new phenomena raise an interesting question for the machine learning community, that is, if the feedforward process can only be approximated, how should we design a more effective gradient update algorithm.

\section{Numerical Experiments\label{sec:experiment}}

\begin{figure}
\begin{center}
\centering
\includegraphics[width=0.45\columnwidth]{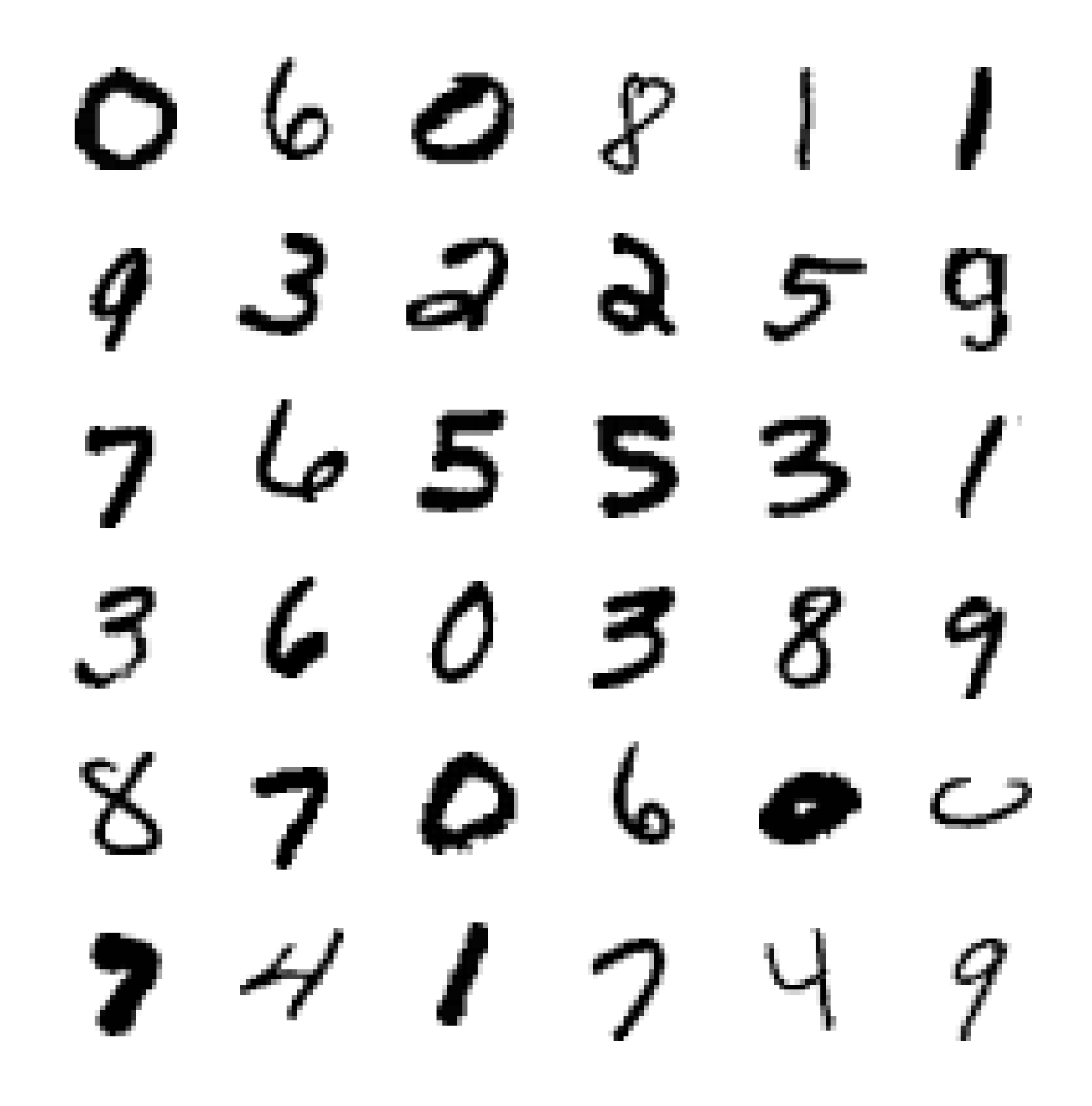}
\includegraphics[width=0.45\columnwidth]{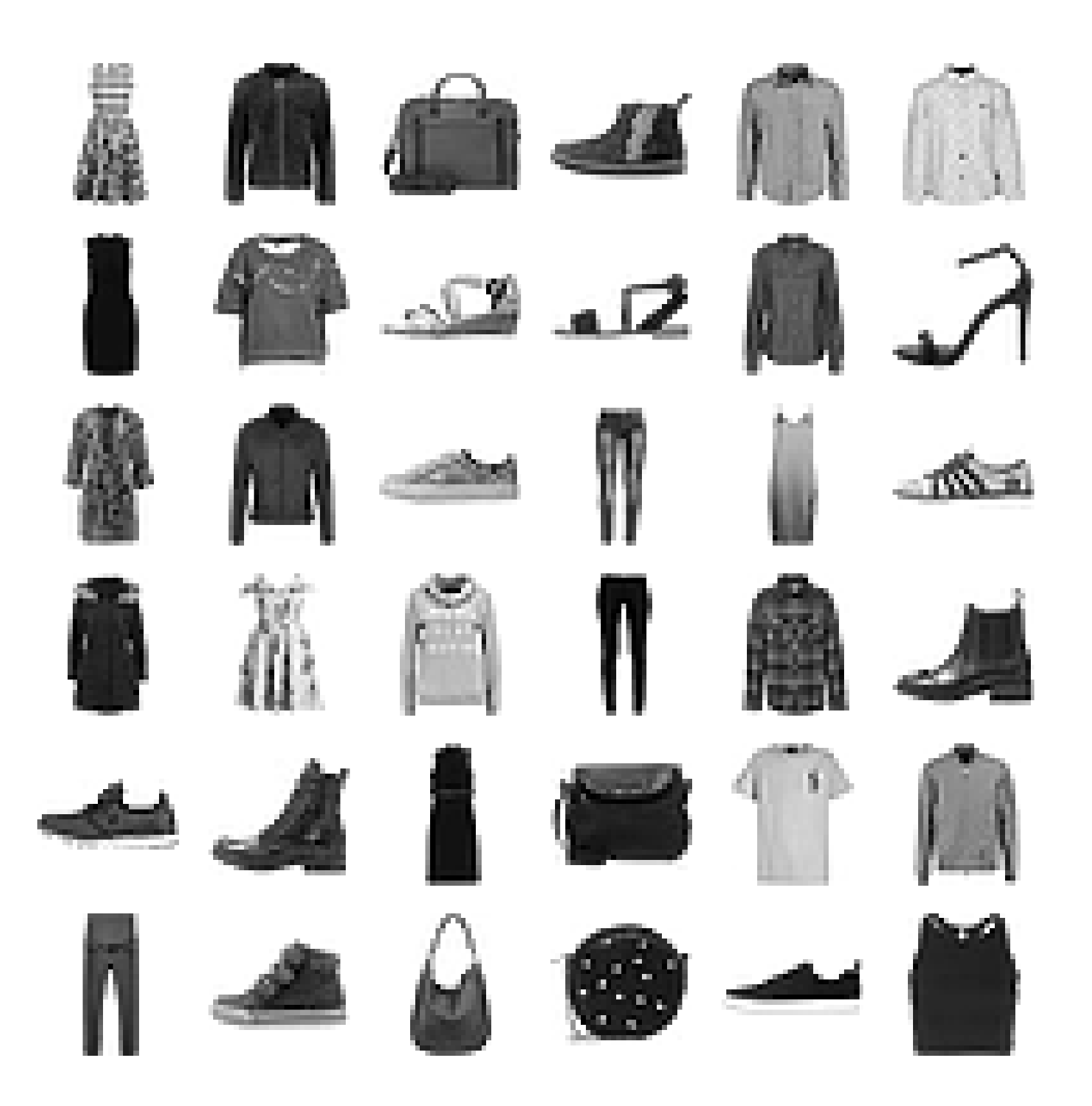}
\caption{Images from the MNIST and Fashion-MNIST dataset.} 
\label{fig:dataset}
\end{center}
\end{figure}

In this section, we conduct numerical experiments to evaluate the expressive power of PEPS models for image classifications. As we have introduced two feature maps for generating input feature tensor to the PEPS classifier, we term the PEPS classifier with the simple product state feature map as \textit{PEPS}, and term the PEPS classifier with the convolution layer feature as \textit{CNN-PEPS}. In PEPS, the simple function feature mapping in Eq.~\ref{eq:feature_map2} is used and further transferred to PEPS tensors with $2\times2$ blocking. This means that for $28\times28$ images, the PEPS would be $14\times14$ with the dimension of physical indices equal to $16$. Each tensor handles the information of pixels within a $2\times2$ square. In practice, we found that constrain the parameters $\theta$ of PEPS models to be positive would significantly improve the stability of optimization. 
The CNN-PEPS shared the $2\times2$ blocking technique and used one layer of CNN as the feature mapping. The CNN layer has 10 convolution filters with size $5\times5$, stride $1$, ReLU activation, and $2\times2$ max pooling. Under this feature map, the positive constraint of the parameter has no significant effect on the optimization result. In both models we set bound dimension of PEPS classifier $\chi = 10$.

To compare with the traditional learning model, we also experimented with fully connected multilayer perceptrons with $784$ input neurons, $n_h$ hidden neurons and $10$ output neurons. The activation function is \textit{softmax} and the cost function is \textit{cross entropy}, the same to the PEPS model. The CNN-MLP has the similar MLP with the same CNN layer of CNN-PEPS used for feature extractions. In our experiments, the best test accuracy is achieved with $n_h = 1000$ for both MLPs.
For fair comparisons, the same hyperparameters are shared by the four models: the learning rate $\alpha = 10^{-4}$, the batch size is $100$,  regularization is set to $0$, weight decay is $0$, and we train $100$ epochs in total.  
To compare with the one-dimensional tensor networks learning model, we also experimented with the MPS model, with parameters set to be exactly the same as in Ref.~\cite{MPSSL}. The code of the MPS model is based on the open-source project\cite{torchmps}.

\subsection{MNIST dataset}
We first test our models using the MNIST dataset~\cite{MNIST}, a simple and standard dataset widely used by many supervised learning models. The MNIST dataset consists of $55,000$ training images, $5,000$ validation images and $10,000$ test images, each image contains $28\times28$ pixels, the content of these images are divided into $10$ classes, corresponding to different handwritten digits from $0$ to $9$. 

\begin{figure}
\begin{center}
\includegraphics[width=\columnwidth]{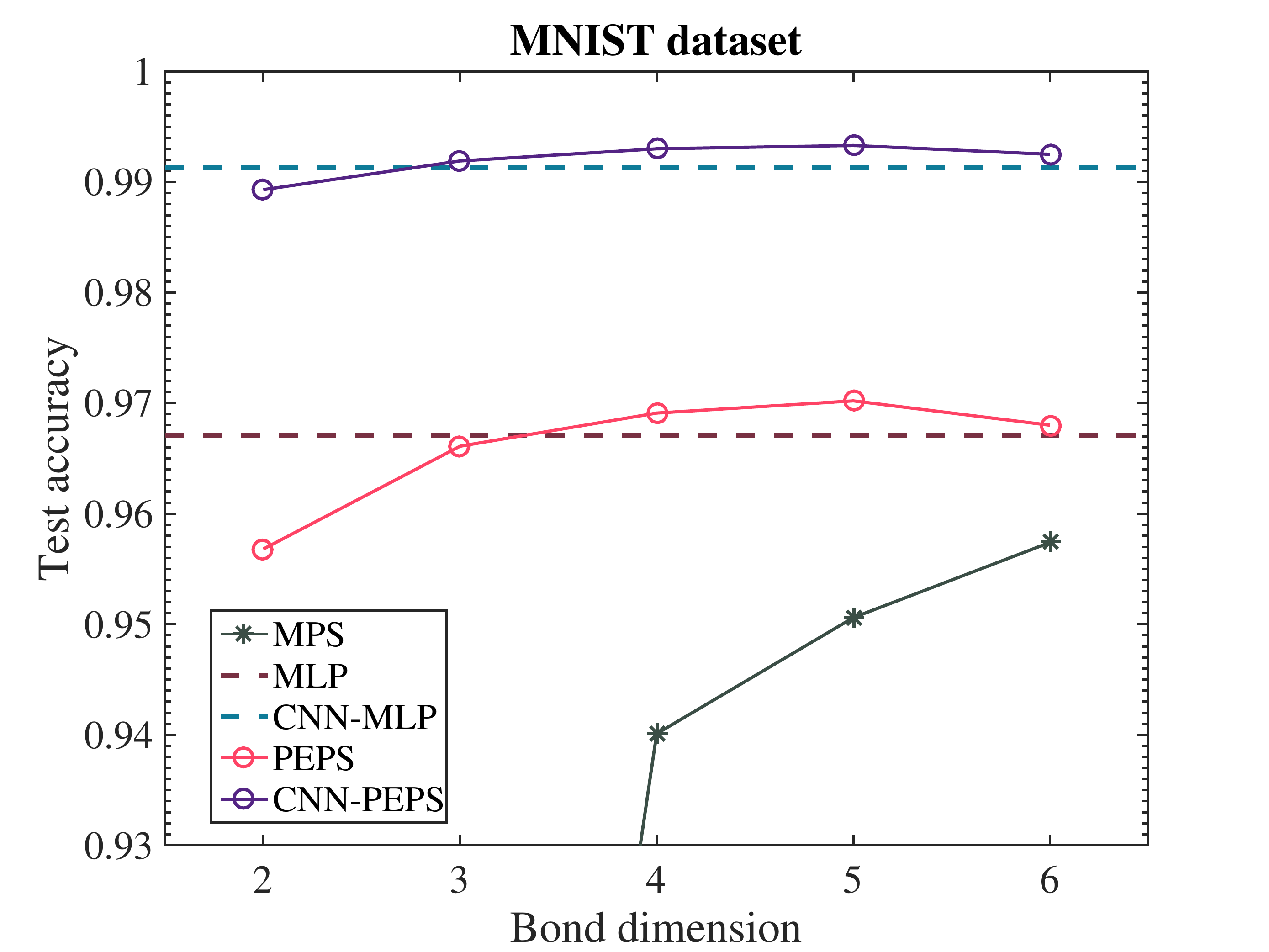}
\caption{Best test set accuracy of different models for MNIST dataset. The dash lines refer to the best results of multilayer perceptrons with $784-1000-10$ neurons. The "CNN" indicates the model applying the convolution feature map in Sec.~\ref{sec:cnn}. Due to the structural prior to images, PEPS models perform significantly better than the one-dimensional MPS model with the same bond dimension. The CNN-PEPS archives the state-of-the-art performance of tensor networks models. Meanwhile, The performance is comparable to the MLP but has fewer parameters. } 
\label{fig:mnistresult}
\end{center}
\end{figure}

As shown in Fig.~\ref{fig:mnistresult}, under the condition of the same bond dimension $D$, the best test accuracy of the PEPS model is significantly better than that of MPS, which reflects the superiority of PEPS tensor networks in modeling images over one-dimensional tensor networks. At the point of $D=5$, the PEPS model achieves its best test set accuracy $97.02\%$. Specifically, one obvious that the PEPS model already performs well when $D$ is small. At the point of $D=2$, the training accuracy of PEPS is already very close to 100\% (99.68\%). With $D=3$, the training accuracy grows to 99.99\%, meaning that only 4 out of 55000 labels are miss predicted.
We also note that with $D = 3$, PEPS and CNN-PEPS give almost the same best test accuracy as MLP and CNN-MLP, while the number of parameters of the PEPS structure is 27.60\% and 6.96\% of the corresponding MLP structure, respectively. These facts imply the potential application of tensor networks in model compression. We also found that the best test accuracy of PEPS structure is stable in a wide range of learning rate($10^{-5}$ to $0.2$) and maximum value of input data($10^{-2}$ to $10^3$), while the best results of MLP easily deteriorated under a smaller perturbations. Moreover, with the bond dimension $D=5$, the CNN-PEPS archives $99.31\%$ test set accuracy of the MNIST dataset, which is the state-of-the-art performance of tensor networks models. Compared with MPS, which archive best test accuracy $99.03\%$ at $D=120$, the good performance at lower $D$ also verifies the inherent low entanglement locality of the natural image dataset itself. This inherent nature of images may be the physical reasons for the success of machine learning models like CNN. Moreover, the PEPS with a small $D$ is beneficial to the hardware implementation of the quantum machine learning model.

\subsection{Fashion-MNIST dataset}
Another dataset we evaluate is the Fashion~MNIST dataset, which includes grayscale photographs of 10 classes of clothing, and is considered as a more challenging dataset than the MNIST dataset.
The test accuracy results of different models are detailed in Table.~\ref{tabel}.
We can see that with the bond dimension $D=5$, the best test accuracy of PEPS-CNN could reach 91.2\%, which is the current state-of-the-art result of the tensor network machine learning model on the Fashion-MNIST dataset. It's also competitive with the AlexNet and XGBoost models, but there is still a clear gap with the most recent advanced convolutional neural network, such as the GoogleNet which employs many convolution layers.
\begin{table}
\centering
\caption{Best test set accuracy of different models for the Fashion-MNIST dataset. Bold are the models calculated in this work. The "CNN" indicates the model applying the convolution feature map in Sec.~\ref{sec:cnn}.}
\label{tabel}
\begin{tabular}{@{}lllll@{}}
\\
\toprule
Model & \multicolumn{4}{l}{Test Acurracy} \\ \midrule
Support Vector Machine~\cite{glasser2018supervised} & \multicolumn{4}{c}{84.1\%}  \\
MPS~\cite{2019arXiv190606329E} & \multicolumn{4}{c}{88.0\%}  \\
\textbf{MLP} & \multicolumn{4}{c}{88.3\%}   \\
\textbf{PEPS}  & \multicolumn{4}{c}{88.3\%}   \\
MPS + TTN~\cite{stoudenmire2018learning} & \multicolumn{4}{c}{89.0\%}  \\
XGBoost~\cite{stoudenmire2018learning} & \multicolumn{4}{c}{89.8\%}  \\
AlexNet~\cite{glasser2018supervised} & \multicolumn{4}{c}{89.9\%}  \\
\textbf{CNN-MLP} & \multicolumn{4}{c}{91.0\%}   \\
\textbf{CNN-PEPS} & \multicolumn{4}{c}{91.2\%}  \\
GoogleNet~\cite{glasser2018supervised} & \multicolumn{4}{c}{93.7\%}  \\ \bottomrule 
\end{tabular}
\end{table}

\section{Conclusions and Discussions \label{sec:discussion}}
We have presented a tensor network model for supervised learning using the PEPS, which directly takes advantage of the two-dimensional feature tensor when compared with MPS and MLP classifiers. We applied the Boundary MPS method to achieve efficient approximate contraction on the feedforward process of the PEPS classifier, and combined the automatic differentiation tensor network technique in the backward propagation process to compute gradients of model parameters.

Using extensive numerical experiments we showed that on both the MNIST and Fashion-MNIST datasets, we obtained the state-of-the-art performance of tensor network learning models on the test accuracy, which illustrates the prior advantages of the two-dimensional tensor networks in modeling natural images. 
Compared with the fully connected neural networks such as MLP, the two-dimensional tensor network structure makes more use of the features of the images itself, and achieves the same or even slightly better results with fewer parameters. More importantly, a machine learning algorithm based on tensor networks has the potential to transform to a quantum machine learning algorithm based on the near-term noisy intermediate-scale quantum circuits. Notably, the PEPS structure has a geometric structure similar to that of several current quantum hardware~\cite{arute_quantum_2019}. Moreover, we have shown that the classical supervised learning algorithm based on PEPS works well even when $D$ is small, so the quantum machine learning model corresponding to this structure could benefit from our result.

However, there are also several issues in the 2D tensor network machine learning models. First, the computation and storage costs of the PEPS model are significantly higher than traditional machine learning models such as neural networks. Second, although the expressive power of the model is likely to be sufficient, the current optimization method may not be suitable for optimizing such complex tensor networks that can only be approximately contracted, resulting in the failure of stable convergence when the model parameters are negative.

Finally, there are many possible ways to improve the current PEPS machine learning model. For example, it benefits from new optimization methods that are more suitable for tensor networks. One could also combine the renormalization group and machine learning in a more practical way by applying the coarse-grained contraction scheme.
Moreover, the unsupervised generative learning model based on two-dimensional tensor networks may be a promising attempt.

\begin{acknowledgments}
We thank E. Miles Stoudenmire, Jing Chen, Jin-Guo Liu, Hai-Jun Liao and Wen-Yuan Liu for inspiring discussions and collaborations. 
S.C. are supported by the National R\&D Program of China (Grant No. 2017YFA0302901) and the National Natural Science Foundation of China (Grants No. 11190024 and No. 11474331).
L.W. is supported by the Ministry of Science and Technology of China under the Grant No. 2016YFA0300603 
and National Natural Science Foundation of China under the Grant No. 11774398. P.Z. is supported by Key Research Program of Frontier Sciences of CAS, Grant No. QYZDB-SSW-SYS032 and Project 11747601 of National Natural Science Foundation of China.
\end{acknowledgments}

\bibliography{peps_sup}

\end{document}